# Context Aware End-to-End Connectivity Management

Jaydip Sen, P. Balamuralidhar, M. Girish Chandra, Harihara S.G., Harish Reddy
Embedded Systems Research Group, Tata Consultancy Services, Bangalore-560066, India,
emails: {jaydip.sen, balamurali.p, m.gchandra, harihara.g, h.reddy}@tcs.com

*Abstract*—In a dynamic heterogeneous environment, such as pervasive and ubiquitous computing, context-aware adaptation is a key concept to meet the varying requirements of different users. Connectivity is an important context source that can be utilized for optimal management of diverse networking resources. Application QoS (Quality of service) is another important issue that should be taken into consideration for design of a context-aware system. This paper presents connectivity from the view point of context awareness, identifies various relevant raw connectivity contexts, and discusses how high-level context information can be abstracted from the raw context information. Further, rich context information is utilized in various policy representation with respect to user profile and preference, application characteristics, device capability, and network QoS conditions. Finally, a context-aware end-to-end evaluation algorithm is presented for adaptive connectivity management in a multi-access wireless network. Unlike the currently existing algorithms, the proposed algorithm takes into account user QoS parameters, and therefore, it is more practical.

## I. INTRODUCTION

The emergence of new wireless broadband networks and diverse miniaturized and personalized networked devices has given rise to a variety of new mobile services in our daily life. Ultimately, these mobile services are executed as we move in different places, at different time, and under different conditions. Hence, these services get a continuously changing information flow from their execution environment. The management of this flow becomes vital for mobile services delivery. This means that a communication paradigm needs to shift from any time and any place into the right time in the right way, as the former may be very intrusive. Context-awareness is a promising way to manage the information flow, as contexts provide information that characterizes user's environment and any object relevant to the interaction between the user and mobile service [1].

For any mobile service the underlying communication is provided by heterogeneous network environment consisting of wireless and wired networks. As depicted in Fig. 1, each network is responsible for a section of the 'end-to-end' communication path between a mobile user and application server placed in a service provider network or a pair of mobile hosts of the same or different network service provider (s).

Mobile connectivity, i.e., persistency of a wireless connection during the act of being mobile and quality of service (QoS) offered by this connection are critical factors for effectiveness of the mobile service delivery. However, in most of the applications a default wireless connection is chosen at a service-design time and assumptions are made regarding its offered QoS. Since a wireless link is usually a bottleneck in the end-to-end communication path, the assumptions regarding its offered QoS imply the assumption for the end-to-end offered QoS. Consequently, current mobile services are delivered with a best-effort (end-to-end) quality, and without any consideration to the mobile user's required QoS [2]. Thus apart from the connectivity, application QoS is also an important criterion on the basis of which application service should be provided. If various connectivity possibilities coexist each offering different QoS, and if the wireless networks are chosen based on their availability and not the application's QoS then the final end-to-end goal may not be achieved. The end-to-end goal of any application in this context is to have uninterrupted service delivery by maintaining a threshold minimum level of performance. Moreover, the user's (application's) requirement may not be static, and thus it may not be possible to satisfy it by the traditional 'best-effort' service. Also the applications may be integrated with learning mechanisms so that they may learn from the end-to-end QoS experienced along the service discovery and life cycles of the applications.

In this paper, context parameters related to connectivity and other QoS issues are discussed. Higher level connectivity context is defined and abstracted from the raw low-level connectivity by integrating other contexts. This context information is utilized to design policies for developing connectivity adaptation in a multi-access network environment. The policies are formulated keeping in mind the end-to-end goal and the QoS requirements of the applications. However, only the end hosts are considered for reconfiguration scenario. The reconfiguration of the intermediate entities (routers, gateways, base-stations) may be considered in future extension of this work. The mechanisms proposed in [3] are largely followed and adapted to achieve the end-to-end goals.

The rest of the paper is organized as follows. Section II discusses some related work on context-aware systems. Section III presents various context information that are relevant in a context-aware system. Section IV depicts a conceptual architecture of a context-aware system and functions of its different components. Policy representation issues and characteristics of adaptive services are discussed in Section V. Section VI presents an algorithm for selection of the most optimum channel of communication between two mobile devices in a context-aware scenario. Finally, Section VII concludes the paper.

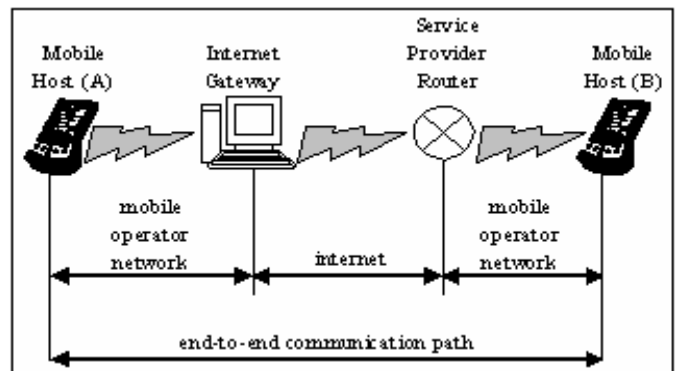

Fig.1. An end-to-end communication path between two mobile hosts

## II. RELATED WORK

In this section, we discuss some related work on context-aware systems. In general, the algorithms for horizontal handoff criteria focus mainly on link quality conditions, e.g., signal strength, SNR (signal to noise ratio), frame error rate, base station workload etc. In vertical handoff, more attention is paid to high-level context information e.g., user preference, cost, application feature, device capacity, bandwidth, etc.

Park et al have proposed an algorithm for seamless handoff between WLAN and CDMA2000 cellular network [9]. Traffic is classified into real-time and non-real time services. The beginning of the handoff is decided by the handoff delay time and throughput according to traffic classes. The parameters for handoff decision include RSS threshold and continuous beacon signal.

Mola has proposed two mechanisms for vertical handoff [10]. The *data-driven mechanism* chooses the network interface that best satisfies the data flow requirement. Parameters that describe data flow include destination, throughput, reliability, and fixed IP source address. The *link-driven mechanism* takes into account two classes of parameters- internal and external. Internal parameters are maximum link speed and reliability; external parameters are billing, cost, and power utilization. The proposed mechanism also considers user preferences.

Aust et al has presented a policy-based Mobile IP handoff decision algorithm [11]. It describes link layer parameters that can be used to control Mobile IP handoffs for seamless mobility. A *generic link layer* is defined above the access networks, and list of parameters is defined. The list includes information about link, environment, neighborhood and link layer management. The algorithm uses the link layer information from the defined generic link layer as input values.

Wang et al have proposed a policy-driven handoff system for heterogeneous wireless networks [14]. The system allows users to express policies on what is the "best" wireless system at any moment, and make tradeoffs among network characteristics and dynamics such as cost, performance and power consumption. The authors have also designed a performance reporting scheme that estimates the current network conditions which serves as input to the policy specification. Given the current 'best' network, the proposed system determines whether the handoff is worthwhile based on the handoff overhead and potential network usage duration.

Zhao et al have proposed two "flow-oriented" mechanisms in the context of Mobile IP to establish a robust and reliable communication link from a mobile host [13]. The first mechanism supports multiple packet deliver methods (such as regular IP, Mobile IP etc.) and adaptively selects the most appropriate one based on the characteristics of each traffic flow. The second mechanism enables a mobile host to make use of multiple network interfaces simultaneously and to control the selection of the most desirable network interfaces for different traffic flows.

Vidales et al have presented a framework named PROTON- a policy-based solution for 4G mobile devices [12]. The objective of the proposed mechanism is to allow users to seamlessly connect to a highly integrated heterogeneous wireless networks. The networking context fragments are grouped into dynamic and static components. Dynamic components include presence, status, signal strength, congestion, flows, velocity, position etc. Static components provide steady data including the profiles of network, application, user, and infrastructure. The proposed framework shows how richer context information can be incorporated into policies and still a lightweight solution for mobile devices can be made.

Balasubramaniam et al have proposed a handover mechanism that is suitable for multimedia applications in pervasive systems [15]. The mechanism is based on handover decision making process that uses context information regarding user devices, user location, network environment, and requested QoS.

Chan et al have suggested a new handover mechanism that combines Mobile IP with fuzzy logic [16]. During the handover initiation, information on the user profile, QoS perceived by the user, and radio link availability are collected. The selection of the most suitable target segment depends mostly on the user profile containing information such as the minimum and maximum cost and the list of segments with the highest and lowest priority. The authors also present some mobility management signaling protocols.

Zhang et al have presented a novel mobility management system for vertical handoff between WWAN and WLAN [17]. The system involves connection manager (CM) that detects changes in the network conditions in a timely and accurate manner, and a virtual connectivity manager (VC) that uses an end-to-end principle to maintain a connection without additional network infrastructure support. A roaming decision maker and a context database act as the interconnection between the CM and the VC. The context information used are user preferences and technical parameters, such as access delay, available bandwidth, and capabilities of the hosts.

Stemm and Katz have introduced a vertical handoff scheme that can handle simultaneous operation of multiple wireless network interfaces [18]. The system allows mobile users to roam in a "wireless overlay network" structure consisting of room-size, building size, and wide-area data networks. The handoff latency of the scheme is bounded by the time it takes for a mobile host to discover that it has moved in or out of a new wireless overlay. The authors also present optimizations to the basic scheme that assumes no knowledge about specific channel characteristics.

Chen et al have proposed a "smart decision model" to perform vertical handoff [19]. The suggested mechanism can identify the "best" network interface and the "best" instant of time for initiation of handoff. A score function is utilized in the model to make smart decision based on various factors such as the properties of the available network interfaces, the system information, and the user preferences.

The CELLO (Cellular Network Optimization based on Mobile Location) project [6] has dealt with the issue of location-based performance assessment of wireless communication networks. The network performance data is stored in a GIS system. However, it considers signal strength as the only context parameter and not the end-to-end QoS.

In the Equanet project [5], a modeling-based performance evaluation method is developed for the end-to-end QoS delivered to a mobile user over heterogeneous network. However, this project only focuses on VOIP and mobile web browsing mobile services and does not consider other services.

The AWARENESS project [7] provides an idea on network resource awareness at the application level. It indicates user context as necessary information for an application to adapt but does not consider the end-to-end goal optimization or resource availability issues.

Compared with these works, the work in this paper is focused on a policy mechanism for channel-based flow level vertical handoff. A rich set of context information is taken into consideration including local host connectivity context, network QoS parameters, application characteristics, as well as user preferences. A modified handoff decision-making algorithm is proposed that takes into account all these context information into account and selects the most optimum channel between two communicating mobile devices.

## III. CONTEXT INFORMATION

Rich contextual information is crucial in ubiquitous applications. In this section, various types of relevant context information are discussed.

In a broad perspective, the context information can be categorized into two types: (i) *Raw context information* and (ii) *Derived context information*. While the former can be obtained directly from the underlying infrastructure or service providers, the latter is deduced from the raw context data.

Both raw context data and derived context information can be stored in a context storage for later retrieval. Any context information can be represented as a 4-tuple: <*entity name, feature, value, time*>. Each entity is identified by its unique name. To describe the context of an entity its features are defined. A feature must have a value. For example, "Joe is at home" is a user context, in which entity is 'Joe', feature is the location and value is Joe's home [8]. Different types of context data are now described in detail below. As we are interested only in connectivity related context data, only the connectivity context data are discussed.

### A. Raw connectivity context

Raw connectivity context (RCC) is the connectivity related contextual information that can be obtained directly from underlying infrastructure, service provider, device driver, platform APIs, applications and users. Essentially, there are three types of such data: *passive* (e.g. availability, condition, ID/name/address, user subscription status or membership status etc.), *active* (e.g. live connection, duration, speed, access, callee, messaging, printing), and related *user events* (e.g. profiling, noting, alarm setting, diverting, calendar, gaming). Passive data is collected by *periodic polling with interruption* approach. Even if a device if equipped with multiple radio interfaces, the user may not be having access (subscription) to all the service provider networks. Therefore, this context information is also important. Some active data may be directly obtained from system logs and others can be gathered on the basis of use. In general, a set of connectivity related events should be first defined, and then monitored and recorded.

Essentially, there are three types of passive and active context data: (i) device context data, (ii) network context data, and (iii) end-to-end context data. In device context, entities include e.g. local/remote end hosts, server, and any terminal network equipment like modem and network interface adapter. In network context, entities include e.g. base station, access point, gateway, proxy, access router, switch, DNS and DHCP server, etc. In end-to-end context, entities are logical end-to-end communication sessions e.g. connections and calls.

Three different methods can be applied to collect network context information: (i) explicit query, (ii) polling, and (iii) event driven approach. In explicit query method, an application can ask for some specific information. Polling may be deployed to keep fetching context at a fixed interval of time and update application behavior accordingly. In event driven method, an application subscribes to some context events and gets informed when those events happen.

For local information when interfaces of the driver program of the device are not sufficient, some probes can be used to detect more precise information. For end-to-end network information, cooperation between local context collection modules collecting those context information is needed These end-to-end context information may include information like available up/down bandwidth and round trip time (RTT), access entities like base station and access point, and core entities such as Home subscriber server (HSS) and router. Most of these entities are maintained by network service provider.

*B. Connectivity Context interpretation*

High-level derived connectivity context is obtained through abstraction and interpretation of the RCCs and application of knowledge-based derivation rules. The interpretation is to be made based on the some knowledge derivation rules. Two types of connectivity context information may be derived: (i) instant context and (ii) predicted context.

(i) *Instant context*: As a mobile device may have multiple network interfaces e.g., GSM, GPRS, WLAN, Bluetooth, etc., evaluation has to be made on all these available network interfaces at the point when a network link for communication is to be established. The purpose of the evaluation process is to determine the optimum network interface (in case of standalone device) or connection (in case of end-to-end application) for channel creation and switching. In determining the optimum connection, the user QoS parameter issues and costs associated with switching should also be brought in.

For determination of the optimum interface, the RCCs that may be used are: interface lists of the host, network interface information (type, speed, status, statistics, access point), signal strength and SNR of the related access points or base stations, charge rates of the related service providers, threshold minimum QoS of an application that can be accepted by the user, the delay and possible interruption in the application for switching the connection at an instant of time and its associated impact on cost and performance on the application.

For evaluation of the *best connection,* remote host connectivity contexts like network interface list and features for each network interface may be considered. Moreover, end-to-end QoS features like RTT and bandwidth, congestion, packet loss rate (judged by number of retransmissions required), signal strength, throughput etc. may also be considered. The difference between the evaluations of the optimum interface vs. the optimum connection lies in whether or not end-to-end raw context information has to be taken into consideration. In the evaluation of the best interface only local host's features are studied, while for the evaluation of the optimum connection, remote host connectivity contexts like network interface list and the features of each list have to be considered.

(ii) *Predicted context*: In predicted context, the prediction can be made on the basis of both time and location. The case of time-oriented prediction is for the scenario that applications may hope to get future connectivity information in order to adjust their behaviors in a proactive manner. For example, a download application may require an idea about the available bandwidth one hour later or when the WLAN interface may be up, so that it can decide about the time to initiate the application. A location-oriented prediction is mainly concerned with remote discovery and prediction of connectivity resource. As an example, a slide show application may want to get the knowledge about the available connectivity of a remote meeting room so that it can determine whether to store presentation locally or to a server.

Time and location are two most important parameters for predicting context information. For time oriented prediction, RCCs are the type and status of each network interface and connection, the base station/access point used by the network interface, the signal strength/SNR of the base station/access point, the bandwidth of the connection, the traffic statistics in the network, speed and direction of movement of the device, base station location, available connectivity resources of a location and time etc. To derive location-oriented prediction, RCCs include the type, speed, status, and the service provider of each network interface.

For prediction of QoS related context information, QoS-context sources should be deployed. It will carry the responsibility of accumulation of logs on delivered end-to-end QoS from different mobile service users and provision of predictions of this QoS (i.e. along the particular trajectory traversed in a particular timeframe) to mobile applications. The concept of QoS-awareness can be made even broader by logging the actually delivered end-to-end QoS information in the mobile service infrastructure and further using it for QoS predictions. It will lead to a proactive QoS-aware and context-aware infrastructure that can improve the delivered QoS of the user application since the context-awareness helps to better capture the user context-dependent QoS requirements.

*C. Connectivity based context derivation*

Raw and derived connectivity context data are used for further deriving high-level application and user specific context to be utilized for maintaining the desired QoS level of user application. Connectivity with other contexts can be used to derive context information of application feasibility and application distribution. Main connectivity context data needed is end-to-end network QoS context like available bandwidth, RTT, and jitter. These information are utilized to arrive at QoS parameters like, minimum throughput, maximum delay, maximum cost, security, privacy requirement etc. In addition, device contexts such as battery, CPU, memory, display, software, operating system, and media coder capacities etc. are also needed. Finally, connectivity context is also an important source in derivation of user context information. Raw context data are processed for the purpose of higher-level context derivation. Higher-level derived user behavioral contexts are: user's presence, location, route, speed, vehicle, object, proximity, surrounding, group, activity etc. One challenge is to develop algorithms for efficient aggregation and transformation of context data to deal with the heterogeneous data format from different devices, in different spaces, belonging to different owners.

IV. SCHEMATIC ARCHITECTURE OF THE SYSTEM

In this section, a conceptual schematic architecture is described that collects context data, refines the collected data and interprets it. Fig. 2 depicts such a system. The context information, as discussed in Section III, are collected by the sensors, processed and stored in the context storage system. The context collector system fetches the processed context data and feeds the context interpreter system. The context interpreter analyzes the context data. The output of the context interpreter allows the mobile device to dynamically select the most optimum connection at any point of time during the life cycle of an application. This adaptation is achieved by maintaining the life cycle of a *channel* which is a logical end-to-end link between physical application components that are located in separate network devices [3]. A '*connection*' is an end-to-end link between two network interfaces of two peer hosts. In case of a multi-access system, there might be a candidate connection set available for a logical channel with each channel using a specific connection to transfer data at a given point of time. The adaptation and reconfiguration subsystem of Fig. 2 performs three major functions: (i) selecting optimum network interface taking into consideration the cost of communication, the QoS requirement of the application and other contexts like signal strength, bandwidth etc, (ii) enabling connection switching taking into account the QoS factors e.g., amount of disruption that can be tolerated in the application, minimum threshold of the application performance (throughput, delay etc.) and (iii) terminating a connection. Initially, an application requests the transport layer to open a new channel and the best available channel is established. At any point of time during the application life-cycle, if an alternative connection is found to be the optimum one, the current connection is seamlessly switched to that. When there is no valid connection available, I/O request of the application is suspended. The application may restart when a suitable interface is available.

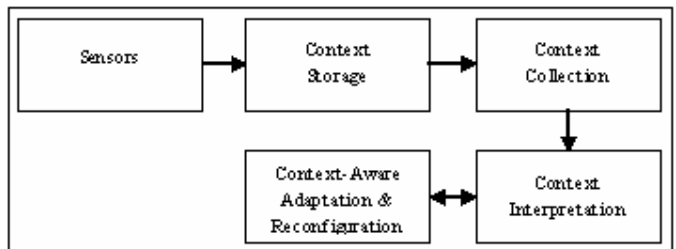

Fig.2. Schematic Architecture of a context-aware system

Every channel has a *traffic class* (TC) associated with it that signifies the class of data transmitted through the channel. A channel may inherit the traffic class form the application that uses the channel. This application-level traffic class can be derived from the QoS of the application like bulk transfer, priority traffic, interactive, responsive, real-time, bandwidth intensive, network control etc. [3].

## V. POLICY REPRESENTATION AND ADAPTATION OF SERVICES

In this section, first a policy representation method is presented and the features of a comprehensive policy representation method are identified. Then, some issues related service adaptation and adaptive services are discussed.

### A. Policy representation

Policies provide a way of managing a collection of resources and employing flexible method of binding resources and policy subjects to the rules in the run time. We propose a 3-tuple representation of a policy ($P$) as follows:

$$P = (TC, RC, EI)$$

The parameter TC is optional and is used to specify the traffic class of the transferred data. Requirement condition RC is also an optional parameter and stands for a set of requirements on the data transmission. The policy evaluation item EI is a mandatory parameter. It is used to construct the policies. There are three types of policies namely *static policy*, *priority policy*, and *weight policy*. Static policy can be represented as: (use/default, type/index, value). The concrete network interface type or index in the device may be explicitly specified or set as default. These are called use policy and default policy respectively. Connection index is the unique identifier of each network interface. Priority policy items are a set of (type/index, value) pairs, with each of them denoting the integer priority value of a particular interface. Weight policy's EI are a set of (factor, weight) pair as: $(f_1,w_1), (f_2,w_2)....(f_n,w_n)$ with $\Sigma w_i=1$. Weight policy enables the selection of connection in run-time.

Three different policy scopes can be defined. A *device level policy* is usually set by the user to express personal preference on the usage of the network connections of the user device as a whole. Each application can also set its own connectivity rules by *application level policy*. A *channel level policy* can be set by an application when a channel is created for data transmission. For each policy definition, there is a setting of channel end type that is used in the evaluation algorithm described in Section VI.

The policy representation is crucial for successful realization and efficient working of any context aware system. An efficient policy representation format should be:

- *Structured*: As policies represent a large number of context information, a structured representation provides for means to filter relevant information effectively. It also eases unambiguous attribute naming as attributes names can be interpreted context sensitively.
- *Interchangeable*: Policies must be interchangeable among different components of the system so that a policy need not be completely retransferred after the change of a single attribute.
- *Composable/Decomposable*: By allowing for profile composition and decomposition, profiles can be stored and maintained in a distributed way. For instance, a default device profile may be stored at the device vendor's web site whereas the deviation from the default is stored in the device itself.
- *Uniform*: A uniform representation of all context policies (device, network, user) eases the interpretation during the process of service mediation and content adaptation.
- *Extensible*: A policy representation format should be adaptable to future extensions.

## B. Service adaptation

Adaptation is one feature of a computing system that can vary the service it provides depending on its input. The computing system, in this context, can be any hardware component or software module at the application level, platform level or infrastructure level. Adaptive services are realized by an effective combination of a set of carefully designed policies and adaptation mechanisms both at the application layer and the platform layer. Adaptive services based on context awareness may be implemented by three different approaches. These three methods are: (i) Application-fully aware method, (ii) Application–unaware method, (iii) Application-aware method.

In application-fully aware method, applications are supposed to be both context-aware and adaptive without any support from the platform including file system, operating system, middleware and network system. In this case, the applications must have all knowledge and access to the context information and adaptation required under different contexts.

In application-unaware method, the applications are completely ignorant about the contexts and adaptation requirements. All these are taken care of the platforms.

In application-aware method, applications together with entities from the platform level cooperate for achieving the goal of adaptive service by taking into account all relevant context information.

It should be noted that when either the platform or the application is of concern, the functionalities of context awareness and adaptation are not necessarily located only in the terminal of the end user, but could be distributed to the whole system including e.g., network gateways and servers and routers.

The allocation of different tasks to different levels- platform or application is a complex design issue. In general, all the common functionalities of different applications should be assigned to the platform level while keeping the application specific issues at the application level. However, from efficiency point of view, if more adaptation functions are assigned to the platform level, better coordination can be done for synchronization of the service components. On the other hand, applications will be more flexible for adaptation if more adaptive tasks are assigned to the application level.

Usually for adaptive systems, the platform level entities are delegated with the responsibilities of collecting, organizing and processing context information shared by all adaptive applications. The operating system and the middleware components are usually used for organizing the context information. The applications can mainly take care of context utilization since the applications know their requirement much better than the lower level system components. The applications can also collect some user interaction contexts for their own use.

For the purpose of adaptation, generally the application should take the final decision about how to adapt itself based on various contexts. The platform should be mostly responsible for execution of the commands as given by the application. The applications guide the adaptation decision at the platform level through the policies.

An adaptive service may be executed through a sequence of stages e.g., adaptation triggering, approach selection and adaptation execution. Adaptation triggering is done by some specific context according to the matching criteria predefined in the system. Then a decision is made about which adaptation approach should be used. Finally, service adaptation is achieved by automatically or manually executing a command and/or changing the external behaviors (and possibly internal states) of an entity that is responsible for providing the service.

Adaptive service can be realized either as a pre-defined adaptation alternative or as a run-time determined solution. For example, to a video streaming service, several versions with different resolutions can be saved at server side beforehand, each of which is for different capabilities of user device and network connection. Another way is to maintain only the original (the highest resolution) version and decide which compression codec should be used in delivery according to the transient context obtained. Actual adaptation mechanisms to be used are closely related to and totally dependent on each specific application type.

Another interesting issue is to decide about the extent to which the user should be involved in the context awareness and adaptation execution. There can be widely varying scenarios in this regard:

(i) The user may wish to be aware of the major events that are happing in the adaptive system, or at least, the user should be bale to know these events if he/she wants.
(ii) The user may even want to manually select the adaptation mechanism of the system himself/herself.
(iii) The user may prefer to be ignorant about the underlying adaptation mechanism and only focus on the quality of service provided by the fully automatic adaptive service.

A simple example of this problem could be the different pre-defined profiles of a mobile phone, e.g., general, silent, meeting, outdoor, customized, etc. A user may prefer his/her mobile phone to automatically change to the 'silent' mode when he/she enters a meeting room and return to the 'general' mode when he/she is out of the room. The user may not prefer this automatic decision making of the phone; instead he/she may prefer an alarm system. The user may also prefer to select the mode explicitly by himself/herself without even an alarm. There are two aspects of this problem: the degree of accuracy of the prediction capability of the system about the user's intention, and the extent of visibility of the adaptation of the system the user prefers.

## VI. EVALUATION ALGORITHMS

In this section, we present an algorithm for channel evaluation in a scenario where two mobile devices directly communicate with each other taking into account various context information that are available. An algorithm for this purpose has been proposed in [3] in which the all the available channels between the communication devices are evaluated, and the best channel is selected. However, this algorithm has a potential drawback. While it chooses the 'best' channel at any instance of time and is guided by some pre-defined policies in the system, it does not take into account the application QoS. In a scenario where the context information changes rapidly, invocation of this algorithm may result in switching of connections too frequently and the delay associated with the connection switch may severely degrade the QoS of the application.

We propose a modified algorithm that selects the 'optimum' channel between two end devices taking into account the application QoS issues and the policies defined in the system. Therefore, the proposed algorithm will not only select the best channel but will also ensure that the application QoS is guaranteed throughout. The algorithm has three stages of execution, which is schematically represented in Fig. 3. The three stages are: (i) Policy traverse, (ii) Cost matrix computation, and (iii) Channel selection. These are discussed in details below.

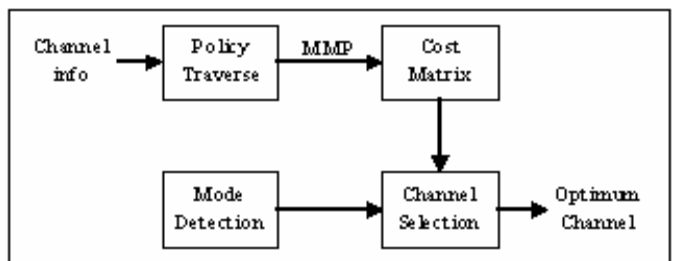

Fig.3. Schematic Architecture of a context-aware system

### A. Policy traverse

The function of the policy traverse module is to search through the related policies in order to find the most matching policy (MMP). MMP is the policy that has the highest matching value (MV). The algorithm for policy traverse returns MMP, which is used for cost matrix computation. The steps of the policy traverse algorithm are as follows:

a. Get relevant channel information including transmission direction, traffic class etc.
b. Calculate MV for each policy staring from the highest priority policy.
c. If one matching policy is found, return it as the MMP.
d. If a set of matching policies is found with only one with the highest MV, return it as the MMP.
e. If a set of matching policies is found with more than one having the highest MV, return the one that as the highest second order priority.
f. If no matching policy is found, go to step b and start with the next level priority.
g. If no matching policy is found, exit and leave the decision to the operating system.

### B. Cost matrix computation

After the MMP is returned, a cost matrix $C$ is generated for the channel according to the selected MMP. Suppose a channel exists between hosts $A$ and $B$. The number of network interfaces of $A$ is $m$ and that of $B$ is $n$. If any end-to-end context factor is used in $A$'s MMP (RE or EI), then host $A$ generates an $m \times n$ matrix $C$. The element $c_{ij}$ of the matrix $C$ is the cost associated when a channel is built upon $A$'s interface $i$ and $B$'s interface $j$. In case only local context factors are used in $A$'s MMP, an $m$ vector is generated. Similarly, host $B$ generates an $n \times m$ matrix or a $n$ vector. A detailed algorithm for the cost matrix computation at host $A$ is as follows:

a. Collect related local end-to-end context information;
b. If MMP's RE and/or EI concerns end-to-end context factors, then for each interface $j$ of host $B$ ( $j \in [1, n]$ ), do step c to generate the column $j$ in the $m \times n$ cost matrix $C$; else do step c ignoring the $j$ for generating the $m$ cost vector $C$;
c. For each network interface $i$ of host $A$, if $i$ ( $i \in [1, m]$ ), is unavailable, or $RE (i, j)$ = false, $c_{ij}$ = infinite, continue with step c for the next interface of host $A$; else set $c_{ij}$ = MAX; add $i$ to the qualified set;
d. For each interface $i$ in the qualified set, do step e to h;
e. If MMP is a *use policy* and $i$ is the specified one, set $c_{ij}$ = 0, exit;
f. If MMP is a *priority policy* and $i$ has a defined priority value, set $c_{ij}$ as $i$'s priority value, continue with step d for the next interface;
g. If MMP is a *weight policy*, calculate $i$'s cost value and assign to $c_{ij}$; continue with step d for the next interface;
h. If MMP is a *default* policy and $i$ is the specified one, set $c_{ij}$ = 0, exit.

In this algorithm, if MMP is a weight policy then cost is the sum of the products of weights and the corresponding values.

### C. Channel selection

After the two sides (i.e. devices) of a channel have both generated their cost matrices, the final decision is made about which connection is the best (minimum cost) and will be used for the channel for the moment. There are two modes of decision-making: master-slave mode and peer-to-peer mode. First the channel end types of both the ends are ascertained. The channel end type is a setting specified by the user or application, as master or salve. Then the decision mode is determined by the XOR (exclusive OR) operation with the channel end types. That is, if the two types are the same, then a peer-to-peer mode is chosen. Otherwise, a master-slave mode is selected. In the following we present our modified algorithm for the master-slave communication type. The modification for the peer-to-peer algorithm will be similar and hence not presented due to space constraints.

Suppose, $C_M$ is the $m \times n$ cost matrix at the master host (MH) and $C_S$ is the $n \times m$ cost matrix at the slave host (SH). The following are the steps of the proposed algorithm:

a. At the MH add all pair $(x, y)$ into candidate set, if $c_{xy}$ has the minimum value in $C_M$. If the pair is unique, return $(x, y)$; else do step b.
b. At the SH by using the exchanged pair $(y, x)$ of each pair $(x, y)$ in the candidate set as the index, search the

corresponding element $c_{yx}$ from $C_S$.

c. At the SH return $(j, i)$ with the index of which the $c_{ji}$ in the $C_S$ has the minimum value.

d. At the MH return the exchanged pair $(i, j)$. The channel is now established between the interface $i$ of $A$ and interface $j$ of $B$.

e. If an event occurs (i.e., the channel conditions change), recompute the matrices $C_M$ and $C_S$ and identify the current best channel. Let the new best channel is between the interfaces $k$ and $l$ of devices $A$ and $B$ respectively.

f. Estimate the time of switching from the channel $(i, j)$ to $(k, l)$. Estimate the impact on the QoS and the associated cost for switching. If the impact on the QoS (delay, throughput, voice and image quality, packet loss etc.) due to switching is not acceptable and the present QoS is above the threshold of acceptable quality, continue with the present channel. Go to step e.

g. If the QoS degradation due to channel switching is acceptable and the cost of communication through the channel $(k, l)$ is also acceptable to the user then switch the channel between $A$ and $B$ from $(i, j)$ to $(k, l)$. Go to step e.

h. If the QoS degradation due to channel switching is acceptable but the cost of communication through the channel $(k, l)$ exceeds what is acceptable to the user, then ask for the user response and switch to the new channel if the user is interested to bear the extra cost; otherwise suspend the channel and wait till a suitable channel comes up. Repeat from step a.

In a real-world scenario, some more parameters can be brought in to make the algorithm more powerful. Before a channel switch or suspension decision is arrived at, an estimation can be made about the temporal aspect of the channel quality. If the algorithm is invoked too frequently due to rapid changes in the channel quality, then the application QoS will degrade instead of improving. Thus before a channel switch is made an estimate should be done about the steadiness of quality of the channel to which the application is switched to. Moreover, the current channel quality may also improve after a small period of degradation in quality. It will be wise to wait for a threshold period of time before invoking the algorithm for channel switch. The time threshold may depend on several factors e.g., the type of the application program, capabilities and types of the end devices, existing channel conditions etc., and its determination can pose a research challenge.

## VII. Conclusions

In this paper, a QoS-aware and context-aware system is presented that supports the execution of mobile applications in a heterogeneous network environment. The raw connectivity context information are identified and higher-level contexts are derived from them. A conceptual architecture is presented for policy based adaptive decision of channel selection based on application QoS issues. Policy representation issues are also discussed. The policy mechanism can be easily extended to include adaptive selection of multiple user devices. The reconfiguration issues of the intermediate entities will be a future scope of work.